\documentclass[12pt,twoside]{article}

\usepackage{taln2000}
\usepackage{times}
\usepackage{graphicx}

\shorttitle{Intermediate Alphabet Reduction \hfill \thepage}
\title{Reduction of Intermediate Alphabets \\
	in Finite-State Transducer Cascades}

\shortauthor{\thepage \hfill Andr\'e Kempe}
\author{Andr\'e Kempe}
\coordonnees{Xerox Research Centre Europe -- Grenoble Laboratory \\
		6 chemin de Maupertuis -- 38240 Meylan -- France \\
		{\small\tt andre.kempe@xrce.xerox.com}}

\newcommand{\aSet}[1]{\{{#1}\}}			

\newcommand{\spc}[1]{\rule{#1}{0mm}}

\def\novalue{\rule[0.6ex]{3ex}{0.1ex}}

\newenvironment{FigText}[1]
{\begin{minipage}{#1}
\small\it}
{\normalsize\rm
\end{minipage}}

\newcounter{LabelValueCounter}
\newcommand{\setLabelValue}[2]			
{\setcounter{LabelValueCounter}{#2} \addtocounter{LabelValueCounter}{-1}
 \refstepcounter{LabelValueCounter} \label{#1}}

\setLabelValue{fst-exm-311}{1}
\setLabelValue{fst-exm-31}{2}
\setLabelValue{fst-exm-312}{3}
\setLabelValue{fst-exm-32}{4}

\begin{document}

\maketitle

\begin{abstract}
  This article describes an algorithm for
  reducing the intermediate alphabets
  in cascades of finite-state transducers (FSTs).
  Although the method modifies the component FSTs,
  there is no change in the overall relation
  described by the whole cascade.
  No additional information or special algorithm,
  that could decelerate the processing of input,
  is required at runtime.
  Two examples from {\it Natural Language Processing}\/
  are used to illustrate the effect of the algorithm
  on the sizes of the FSTs and their alphabets.
  With some FSTs the number of arcs and symbols shrank considerably.
\end{abstract}



\section{Introduction}

This article describes an algorithm for
reducing the intermediate alphabet
occurring in the middle of a pair of finite-state transducers (FSTs)
that operate in a cascade,
i.e., where the first FST
maps an input string to a number of intermediate strings
and the second maps those to a number of output strings.
With longer cascades,
the algorithm can be applied pair-wise to all FSTs.
Although the method modifies the component FSTs
and component relations that they describe,
there is no change in the overall relation
described by the whole cascade.
No additional information or special algorithm,
that could decelerate the processing of input,
is required at runtime.

{\it Intermediate alphabet reduction}\/
can be beneficial for many practical applications
that use FST cascades.
In {\it Natural Language Processing}\/,
FSTs are used for many basic steps
\cite{karttunen-xfst,mohri97},
such as phonological \cite{kaplan-kay}
and morphological analysis \cite{kosken83},
part-of speech disambiguation
  \cite{roche-brillfst,kempe-acl,kempe-nemlap},
spelling correction \cite{oflazer96},
and shallow parsing \cite{kosken92,ait}.
Some of these applications, such as shallow parsing, use FST cascades.
Others could jointly be used in a cascade.
In these cases,
the proposed method can reduce the sizes of the FSTs.

The described algorithm has been implemented.

\subsection{Conventions \label{ss-conven}}
The conventions below are followed in this article.

{\bf Examples and figures:}
Every example is shown in one or more figures.
The first figure usually shows the original network or cascade.
Possible following figures show modified forms of the same example.
For example,
Example~\ref{fst-exm-311} is shown in Figure~\ref{f-exm311-P}
and Figure~\ref{f-exm311-Pr}.

{\bf Finite-state graphs:}
Every network has one initial state, labeled with number~0,
and one or more final states marked by double circles.
The initial state can also be final.
All other state numbers and all arc numbers have no meaning
for the network but are just used to reference a state or an arc
from within the text.
An arc with {\it n} labels designates a set of {\it n} arcs
with one label each
that all have the same source and destination.
In a symbol pair occurring as an arc label,
the first symbol is the input and the second the output symbol.
For example, in the symbol pair {\tt a:b},
{\tt a} is the input and {\tt b} the output symbol.
Simple (i.e. unpaired) symbols occurring as an arc label,
represent identity pairs.
For example, {\tt a} means {\tt a:a}.
The question mark, ``?'',
denotes (and matches) all unknown symbols, i.e.,
all symbols outside the alphabet of the network.

{\bf Input and output side:}
Although FSTs are inherently bidirectional,
they are often intended to be used in a given direction.
The proposed algorithm is performed wrt. the direction of application.
In this article, the two sides (or tapes or levels)
of an FST are referred to as {\it input side} and {\it output side}\/.


\section{Previous Work}

The below described algorithm of intermediate alphabet reduction
is related to the idea of {\it label set reduction}\/
\cite{kosken83,karttunen-TwoLevel}.
The later is applied to a single FST or automaton.
It groups all arc labels into equivalence classes,
regardless whether these are atomic labels (e.g. ``{\tt a}''),
identity pairs (e.g. ``{\tt a:a}''),
or non-identity pairs (e.g. ``{\tt a:x}'').
Labels that always co-occur
on arcs with the same source and destination state,
are put into the same equivalence class.
One label is then selected from every class to represent the class.
All other labels are removed from the alphabet,
and the corresponding arcs are removed from the network,
which can lead to a considerable size reduction.
Label set reduction is reversible,
based on the information about the equivalence classes.
At runtime,
this information is required together with a special algorithm
to interpret every label in the network
as the set of labels in the corresponding equivalence class.
For example,
if the label {\tt a} represents the class
\aSet{{\tt a},~ {\tt a:b},~ {\tt b},~ {\tt c:z}}
then it must map {\tt c}, occurring at the input, to {\tt z}.


\section{Reduction of Intermediate Alphabets}

The algorithm of intermediate alphabet reduction
is applied to a pair of FSTs that operate in a cascade
rather than to a single FST.
It reduces the intermediate alphabet between the two FSTs
without necessarily reducing the label sets of the FSTs.
With longer cascades,
the algorithm can be applied pair-wise to all FSTs.
Although the component FSTs and component relations
that they describe are (irreversibly) modified,
there is no change in the overall relation
described by the whole cascade.
No additional information or special algorithm,
that could decrease the processing speed,
is required at runtime.
The fact that every intermediate symbol
actually represents a set of (one or more) symbols,
can be neglected at that point.
Every symbol will be considered at runtime
just as itself.

\subsection{Alphabet Reduction in Transducer Pairs}

We will first describe the algorithm for an FST pair
where the two FSTs, $T_1$ and $T_2$, operate in a cascade,
i.e., $T_1$ maps an input string to a number of intermediate strings
which, in turn, are mapped by $T_2$ to a number of output strings.

\begin{figure}[htbp]
\begin{center}
\includegraphics[scale=0.50,angle=0]{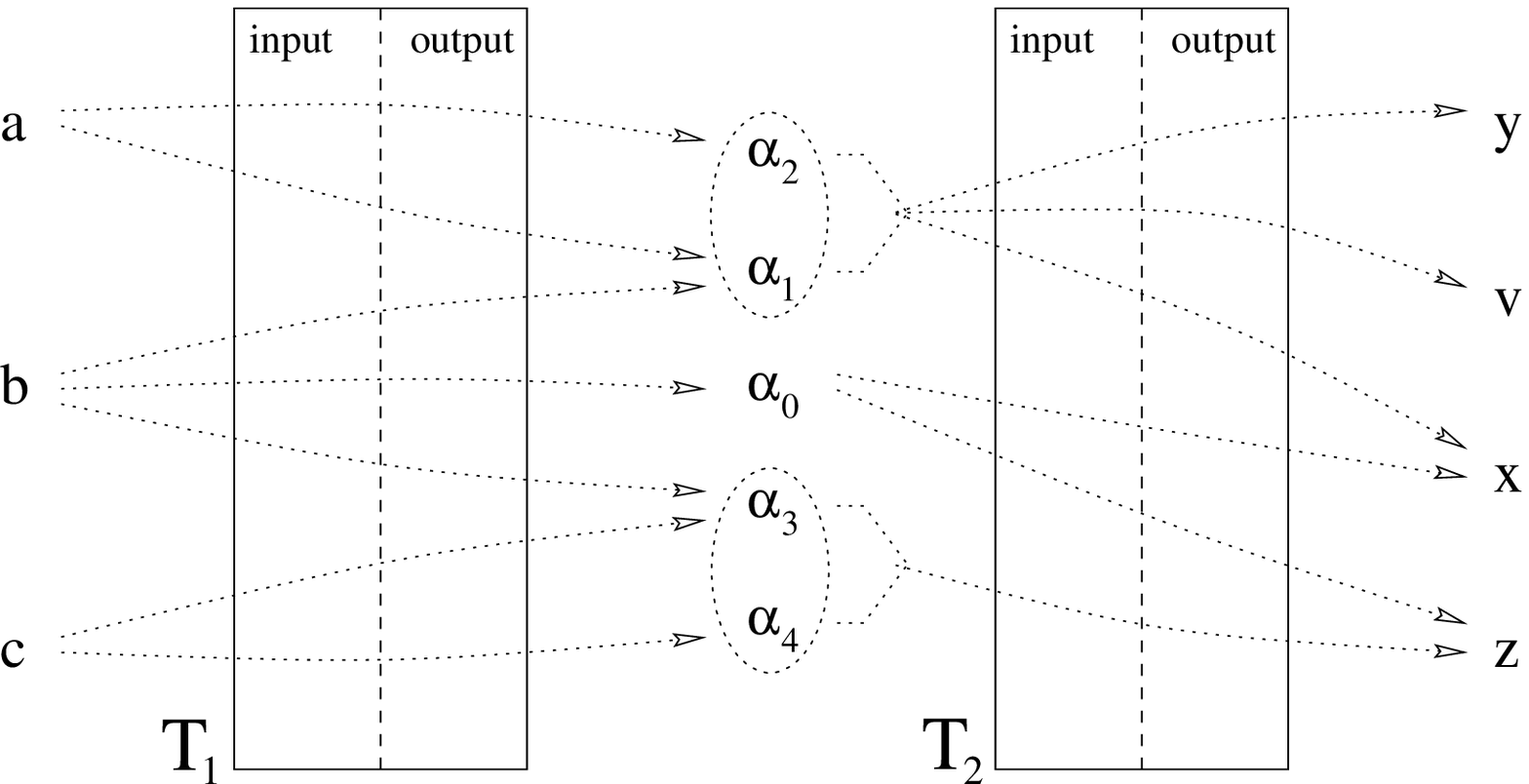}

\vspace{2ex}
{\small
{\it Equivalence classes:}
\aSet{$\alpha_0$},
\aSet{$\alpha_1, \alpha_2$},
\aSet{$\alpha_3, \alpha_4$} }

\vspace{0mm}
\caption{
Transducer pair
  (Example~\ref{fst-exm-311})  \label{f-exm311-P}}
\end{center}
\end{figure}

\begin{figure}[htbp]
\begin{center}
\includegraphics[scale=0.50,angle=0]{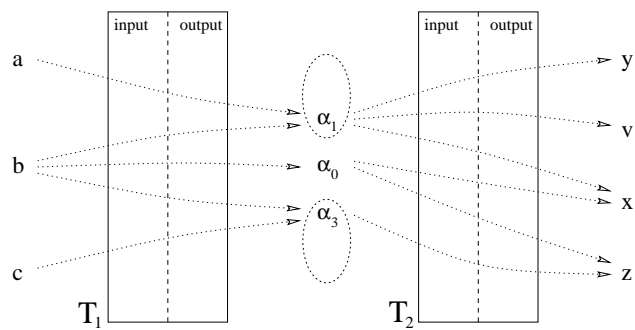}

\vspace{0mm}
\caption{
Transducer pair with reduced intermediate alphabet
  (Example~\ref{fst-exm-311})  \label{f-exm311-Pr}}
\end{center}
\end{figure}

Example~\ref{fst-exm-311}
shows an FST pair
and part of its input, intermediate, and output alphabet
(Fig.~\ref{f-exm311-P}).
Suppose,
both intermediate symbols $\alpha_1$ and $\alpha_2$
are always mapped to the same output symbol
which can be {\tt y}, {\tt v}, or {\tt x}
depending on the context.
This means,
$\alpha_1$ and $\alpha_2$ constitute an equivalence class.
There may be another class formed by $\alpha_3$ and $\alpha_4$.
If we are not interested in the actual intermediate symbols
but only in the final output,
we can select one member symbol of every class to represent the class,
and replace all other symbols by the representative of their class.
In Example~\ref{fst-exm-311},
this means that $\alpha_2$ is replaced by $\alpha_1$,
and $\alpha_4$ by $\alpha_3$
(Fig.~\ref{f-exm311-Pr}).

\begin{figure}[htbp]
\begin{center}
\includegraphics[scale=0.50,angle=0]{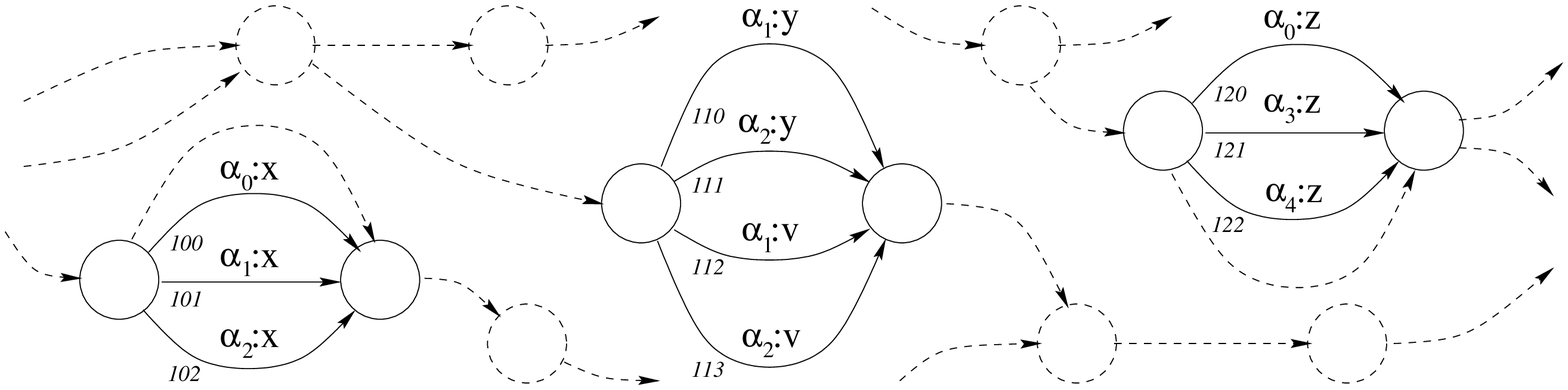}

\vspace{2ex}
{\small
{\it Equivalence classes:}
\aSet{$\underline{\alpha_0}$},
\aSet{$\underline{\alpha_1}, \alpha_2$},
\aSet{$\underline{\alpha_3}, \alpha_4$} }

\vspace{0mm}
\caption{
Second transducer of a pair
  (Example~\ref{fst-exm-31})  \label{f-exm31-f2}}
\end{center}
\end{figure}

\begin{figure}[htbp]
\begin{center}
\includegraphics[scale=0.50,angle=0]{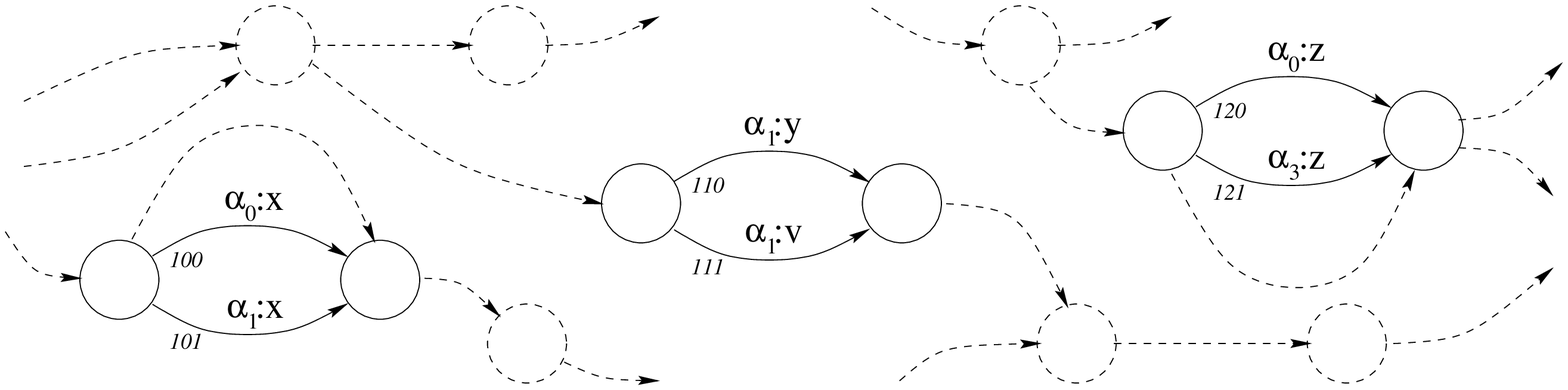}

\vspace{0mm}
\caption{
Second transducer of a pair with reduced intermediate alphabet
  (Example~\ref{fst-exm-31})  \label{f-exm31-f2r}}
\end{center}
\end{figure}

The algorithm works as follows.
First,
equivalence classes are constituted
among the input symbols of $T_2$
(Fig.~\ref{f-exm31-f2}).\footnote{
In Figure~\ref{f-exm31-f2}, \ref{f-exm31-f2r}, and~\ref{f-exm32}
only transitions and states that are relevant for the current purpose
are represented by solid arcs and circles,
and all the others by dashed arcs and circles.}
For this purpose,
all symbols are put initially into one single class
which is then more and more partitioned as the arcs of $T_2$ are inspected.
The construction of equivalence classes terminates
either when all arcs have been inspected
or when the maximal partitioning (into singleton classes) is reached.

Two symbols, $\alpha_i$ and $\alpha_j$, are considered equivalent
if for every arc with $\alpha_i$ as input symbol,
there is another arc with $\alpha_j$ as input symbol and vice versa,
such that both arcs have the same source and destination state
and the same output symbol.
In Example~\ref{fst-exm-31},
we constitute the equivalence classes
\aSet{$\alpha_0$}, \aSet{$\alpha_1, \alpha_2$}, and \aSet{$\alpha_3, \alpha_4$}
(Fig.~\ref{f-exm31-f2}).
Here, $\alpha_0$ constitutes a class on its own
because it first co-occurs with $\alpha_1$ and $\alpha_2$
in the arc set \aSet{\it 100, 101, 102},
and later with $\alpha_3$ and $\alpha_4$ in the arc set \aSet{\it 120, 121, 122}.

Subsequently,
all occurrences of intermediate symbols are replaced
by the representative of their class.
In Example~\ref{fst-exm-31},
we selected the first member of each class as its representative.
The replacement must be performed both
on the output side of $T_1$ and on the input side of $T_2$.
Figure~\ref{f-exm31-f2r}
shows the effect of this replacement on $T_2$ in Example~\ref{fst-exm-31}.

The algorithm can be applied to any pair of FSTs:
They can be ambiguous or they can contain
``{\tt ?}'', the unknown symbol,
or $\epsilon$, the empty string,
or even $\epsilon$-loops.
Although $\epsilon$ and {\tt ?} cannot be merged with other symbols,
this does not represent a restriction for the type of the FSTs.

\subsection{Alphabet Reduction in Transducer Cascades}

\begin{figure}[htbp]
\begin{center}
\includegraphics[scale=0.50,angle=0]{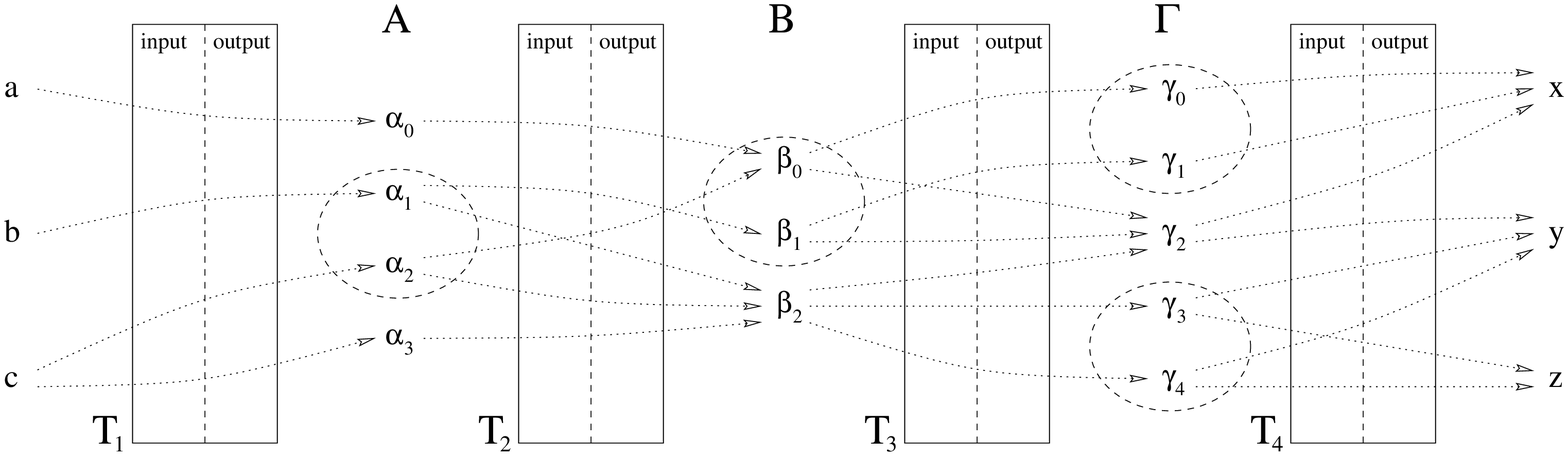}

\vspace{2ex}
{\small
{\it Equivalence classes in different alphabets:} \\
$A$:\aSet{%
\aSet{$\alpha_0$},
\aSet{$\alpha_1, \alpha_2$},
\aSet{$\alpha_3$}}~,~~
$B$:\aSet{%
\aSet{$\beta_0, \beta_1$},
\aSet{$\beta_2$}}~,~~
${\mit\Gamma}$:\aSet{%
\aSet{$\gamma_0, \gamma_1$},
\aSet{$\gamma_2$},
\aSet{$\gamma_3, \gamma_4$}}
}

\vspace{0mm}
\caption{
Transducer cascade
  (Example~\ref{fst-exm-312})  \label{f-exm312-C}}
\end{center}
\end{figure}

\begin{figure}[htbp]
\begin{center}
\includegraphics[scale=0.50,angle=0]{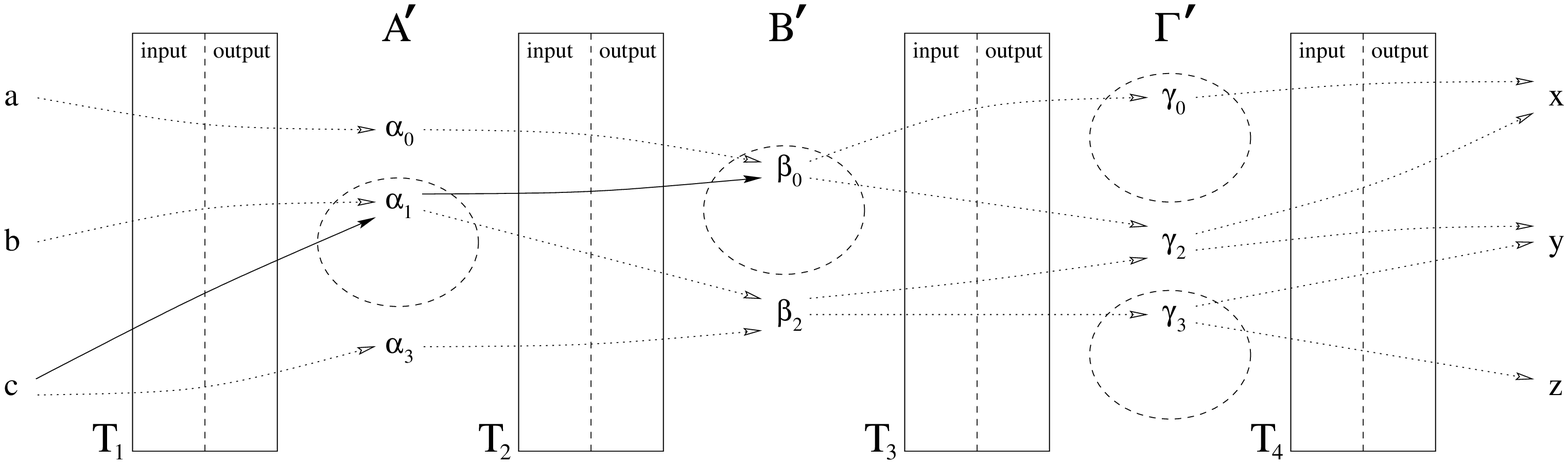}

\vspace{4mm}
\begin{FigText}{60mm}
Solid lines mark modified relations.
\end{FigText}

\vspace{0mm}
\caption{
Transducer cascade with reduced intermediate alphabets
  (Example~\ref{fst-exm-312})  \label{f-exm312-Cr}}
\end{center}
\end{figure}

In a cascade,
intermediate alphabet reduction can be applied pair-wise to all FSTs
starting from the end of the cascade.
Example~\ref{fst-exm-312}
shows a cascade of four FSTs with the intermediate alphabets
$A$, $B$, and ${\mit\Gamma}$
(Fig.~\ref{f-exm312-C}).
The reduction is first applied to the last intermediate alphabet,
${\mit\Gamma}$, between $T_3$ and $T_4$.
Suppose,
there are three equivalence classes in ${\mit\Gamma}$, namely
\aSet{$\gamma_0, \gamma_1$},
\aSet{$\gamma_2$},
and \aSet{$\gamma_3, \gamma_4$}.
According to the above method,
the class \aSet{$\gamma_0, \gamma_1$} can be represented by $\gamma_0$,
\aSet{$\gamma_2$} by $\gamma_2$,
and \aSet{$\gamma_3, \gamma_4$} by $\gamma_3$.
This means,
all occurrences of $\gamma_1$ are replaced by $\gamma_0$
and all occurrences of $\gamma_4$ by $\gamma_3$,
both on the output side of $T_3$ and on the input side of $T_4$
(Fig.~\ref{f-exm312-C},~\ref{f-exm312-Cr}).
Consequently,
$\beta_1$ in the preceding alphabet, $B$,
will now be mapped to $\gamma_0$ instead of $\gamma_1$,
and $\beta_2$ to $\gamma_3$ instead of $\gamma_4$.
The latter mapping actually exists already.

Subsequently,
the preceding intermediate alphabet, $B$, is reduced.
It may contain two equivalence classes,
\aSet{$\beta_0, \beta_1$} and \aSet{$\beta_2$}.
Both members of \aSet{$\beta_0, \beta_1$}
are at present mapped either to $\gamma_0$
(previously \aSet{$\gamma_0, \gamma_1$})
or to $\gamma_2$, depending on the context.
Note,
the class \aSet{$\beta_0, \beta_1$} can only be constituted
if the alphabet ${\mit\Gamma}$ has been previously reduced
and \aSet{$\gamma_0, \gamma_1$} was replaced by $\gamma_0$.
Finally,
we reduce the intermediate alphabet $A$,
based on its equivalence classes
\aSet{$\alpha_0$},
\aSet{$\alpha_1, \alpha_2$},
and \aSet{$\alpha_3$}
(Fig.~\ref{f-exm312-Cr}).

\begin{figure}[htbp]
\begin{center}
\begin{minipage}{60mm}
\includegraphics[scale=0.50,angle=0]{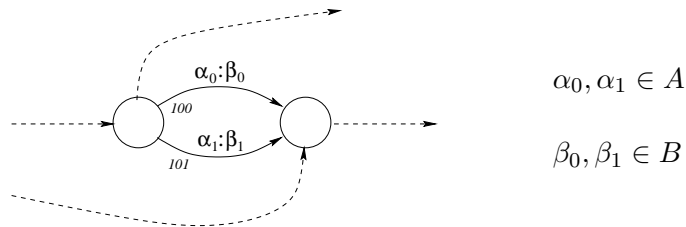}
\end{minipage}
\spc{10mm}
\begin{minipage}{20mm}
{\small
$\alpha_0, \alpha_1 \in A$ \\

$\beta_0, \beta_1 \in B$ }
\end{minipage}

\vspace{0mm}
\caption{
Transducer inside a cascade
  (Example~\ref{fst-exm-32})  \label{f-exm32}}
\end{center}
\end{figure}

The best overall reduction of the cascade is guarantied
if the algorithm starts with the last intermediate alphabet,
and processes all alphabets in reverse order.
This is for the following reason:
Suppose,
an FST inside a cascade maps $\alpha_0$ to $\beta_0$
and $\alpha_1$ to $\beta_1$
on two arcs with the same source and destination state
(Fig.~\ref{f-exm32}).
If the alphabet $B$ is processed first
and $\beta_0$ and $\beta_1$ are reduced to one symbol
(e.g. $\beta_0$)
then $\alpha_0$ and $\alpha_1$ will have equal output symbols
and may therefore become reducible to one symbol
(e.g. $\alpha_0$),
depending on their other output symbols in this FST.
If $\beta_0$ and $\beta_1$ cannot be reduced
then subsequently $\alpha_0$ and $\alpha_1$ cannot be either
because they have different output symbols,
at least in this case.
This means,
reducing $B$ first is either beneficial or neutral
for the subsequent reduction of $A$.
Not reducing $B$ first is always neutral for $A$.
Therefore,
the intermediate alphabets of a cascade should be reduced
all in reverse order.

\section{Complexity}

The running time complexity of constructing equivalence classes
is in the general case quadratic to the number of outgoing
arcs on every state of the second FST, $T_2$, of a pair:
\begin{equation}
O( \; \sum_{q \in Q_2} |E(q)|^2 \; )
\end{equation}

\noindent
where $|E(q)|$ is the number of outgoing arcs of state $q$.
This is because in general every arc of $q$ must be compared
to every other arc of $q$.
In the extremely unlikely worst case
when all arcs have the same source state,
this complexity is quadratic to the total number of arcs in $T_2$~:
\begin{equation}
O( \; |E_2|^2 \; )
\end{equation}

\noindent
The running time complexity of replacing symbols by the representatives
of their equivalence class is in any case linear
to the total number of arcs in $T_1$ and $T_2$~:
\begin{equation}
O( \; |E_1| + |E_2| \; )
\end{equation}

The space complexity of constructing equivalence classes
is linear to the size of the intermediate alphabet 
that is to be reduced:
\begin{equation}
O( \; |\Sigma_{\rm mid}| \; )
\end{equation}

\noindent
The replacement of symbols by the representatives
of their class requires no additional memory space.


\section{Evaluation}

Table~\ref{t-evalu1}
shows the effect of intermediate alphabet reduction
on the sizes of FSTs and their alphabets in a cascade.
The twelve FSTs are used for shallow parsing of French text,
e.g., for the marking of noun phrases, clauses, and syntactic relations
such as subject, inverted subject, or object
\cite{ait}.
An input string to the cascade consists of surface word forms,
lemmas, and part-of-speech tags of a whole sentence.
The output is ambiguous and consists alternatively
of a parse (only one per sentence)
or of a syntactic relation such as ``{\tt SUBJ(Jean,mange)}''
(several per sentence).

\begin{table}[th]\tabcolsep1ex	
\begin{center}
\begin{math}
\begin{tabular}{|r|*{2}{rrl rr|}} \hline
	& \multicolumn{5}{c|}{originally}
	& \multicolumn{5}{c|}{after reduction}  \\
FST	& \#states & \#arcs  &\spc{1ex}& \multicolumn{2}{c|}{\#symbols}
	& \#states & \#arcs  &\spc{1ex}& \multicolumn{2}{c|}{\#symbols} \\
	& 	   &	     & & input  & output
	& 	   &	     & & input  & output    \\ \hline
 1  &
 10~487	&   404~903	&	& 138		& 137	&
 10~487	&   404~903	&	& 138		& 137	\\
 2  &
    604	&    28~569	&	& 136		& 135	&
    604	&    28~569	&	& 136		& 135	\\
 3  &
 27~704	&   225~215	&	&  11		&  10	&
 27~704	&   225~215	&	&  11		&  10	\\
 4  &
  3~613	&    61~259	&	&  23		&  23	&
  3~613	&    61~259	&	&  23		&  23	\\ \hline
 5  &
  1~276	&   128~222	&	& 146		& 142	&
  1~276	&{\bf 124~754}	&	&{\bf 143}	& 142	\\
 6  &
  3~293	&    29~079	&	&  12		&  12	&
  3~293	&    29~079	&	&  12		&  12	\\
 7  &
  5~544	&   166~704	&	&  34		&  33	&
  5~544	&{\bf 90~024}	&	&{\bf 19}	&{\bf 18}	\\
 8  &
    396	&    19~008	&	&  48		&  48	&
    396	&{\bf 12~276}	&	&{\bf 31}	&{\bf 31}	\\ \hline
 9  &
  7~009	&   370~419	&	&  54		&  54	&
  7~009	&{\bf 204~411}	&	&{\bf 30}	&{\bf 27}	\\
10  &
  6~033	& 1~156~053	&	& 158		& 158	&
  6~033	&{\bf 498~506}	&	&{\bf 66}	&{\bf 65}	\\
11  &
    573	&   114~328	&	& 171		& 171	&
    573	&{\bf  52~801}	&	&{\bf 78}	&{\bf 45}	\\
12  &
      2	&       288	&	& 144		&  16	&
      2	&{\bf      34}	&	&{\bf 17}	&  16	\\ \hline
total &
 66~534	& 2~704~047	&	& \novalue	& \novalue	&
 66~534	&{\bf 1~731~831}&	& \novalue	& \novalue	\\ \hline
\end{tabular}
\end{math}

\vspace{4mm}
\begin{FigText}{60mm}
Bold numbers denote modifications.
\end{FigText}

\vspace{0mm}
\caption{
Sizes of transducers and of their alphabets in a cascade
  \label{t-evalu1}}
\end{center}
\end{table}

With some FSTs the reduction was considerable.
For example,
the number of arcs of the largest FST~(\#10)
was reduced from 1~156~053 to 498~506.
In some other cases no reduction was possible.
For the whole cascade the number of arcs
has shrunk from 2~704~047 to 1~731~831
which represents a reduction of 36\%.

Each of these FSTs has its own alphabet which can contain ``{\tt ?}''
that matches any symbol that is not explicitly mentioned
in the alphabet of the FST.
Therefore,
an FST can have a different number of output symbols
than the following FST has input symbols
because a given symbol may be part of the unknow alphabet in one FST
but not in the other.

\begin{table}[th]\tabcolsep1ex	
\begin{center}
\begin{math}
\begin{tabular}{|cc|*{2}{rrl rr|}} \hline
\multicolumn{2}{|c|}{Lexicon}
	& \multicolumn{5}{c|}{originally}
	& \multicolumn{5}{c|}{after reduction}  \\
\multicolumn{2}{|c|}{FST}
	& \#states & \#arcs  &\spc{1ex}& \multicolumn{2}{c|}{\#symbols}
	& \#states & \#arcs  &\spc{1ex}& \multicolumn{2}{c|}{\#symbols} \\
\multicolumn{2}{|c|}{~}
	& 	   &	     & & input  & output
	& 	   &	     & & input  & output    \\ \hline
 E	& $T$
	& 62~120	& 156~757	&	&    224	&    298
	& \novalue	& \novalue	&	& \novalue	& \novalue	\\
 n	& $T_1$
	& 75~900	& 191~687	&	&    224	&  8~874
	&{\bf 73~780}	&{\bf 181~829}	&	&    224	&{\bf  3~042}	\\
 g	& $T_2$
	& 16~748	&  36~737	&	&  8~729	&    153
	& 16~748	&{\bf  24~483}	&	&{\bf  2~897}	&    153	\\
 	& $T_1\!+\!T_2$
	& 92~648	& 228~424	&	& \novalue	& \novalue
	&{\bf 90~528}	&{\bf 206~312}	&	& \novalue	& \novalue	\\ \hline
 F	& $T$
	& 55~725	& 130~139	&	&    224	&    269
	& \novalue	& \novalue	&	& \novalue	& \novalue	\\
 r	& $T_1$
	& 61~467	& 164~819	&	&    224	& 11~420
	&{\bf 59~697}	&{\bf 159~187}	&	&    224	&{\bf  2~680}	\\
 e	& $T_2$
	&  6~814	&  59~587	&	& 11~241	&     90
	&  6~814	&{\bf  42~007}	&	&{\bf  2~501}	&     90	\\
 	& $T_1\!+\!T_2$
	& 68~281	& 224~406	&	& \novalue	& \novalue
	&{\bf 66~511}	&{\bf 201~194}	&	& \novalue	& \novalue	\\ \hline
\end{tabular}
\end{math}

\vspace{4mm}
\begin{FigText}{120mm}
$T$~....~original FST,~~
$T_1$, $T_2$, $T_1\!\!+\!T_2$~....~first and second FST from factorization
and sum of both.~~
Bold numbers denote modifications.
\end{FigText}

\vspace{0mm}
\caption{
Sizes of transducers and of their alphabets in factorization
  \label{t-evalu2}}
\end{center}
\end{table}

Intermediate alphabet reduction can also be useful with
the factorization of a finitely ambiguous FST into two FSTs,
$T_1$ and $T_2$, that operate in a cascade
\cite{kempe-ciaa}.
Here,
$T_1$ maps any input symbol that originally has ambiguous output,
to a unique intermediate symbol
which is then mapped by $T_2$ to a number of different output symbols.
$T_1$ is unambiguous.
$T_2$ retains the ambiguity of the original FST,
but it is {\it fail-safe}\/ wrt. $T_1$.
This means,
the application of $T_2$ to the output of $T_1$ never leads to a state
that does not provide a transition for the next input symbol,
and always terminates in a final state.
This factorization can create many redundant intermediate symbols,
but their number can be reduced by the above algorithm.
Table~\ref{t-evalu2}
shows the effect of alphabet reduction
on FSTs resulting from the factorization of an English and a French lexicon.

Since $T_1$ is unambiguous
it can be further factorized into a left- and a right-sequential FST
that jointly represent a bimachine
\cite{schbrg61}.
The intermediate alphabet of a bimachine, however,
can be limited to the necessary minimum
already during factorization
\cite{elgot,berstel,reutenauer,roche-facto},
so that the above algorithm is of no use in this case.


\section{Conclusion}

The article described an algorithm for
reducing the intermediate alphabets in an FST cascade.
The method modifies the component FSTs
but not the overall relation
described by the whole cascade.
The actual benefit consists in a reduction of the sizes of the FSTs.

Two examples from Natural Language Processing
have been used to illustrate the effect of the alphabet reduction
on the sizes of FSTs and their alphabets,
namely a cascade for shallow parsing of French text
and FST pairs resulting from the factorization of lexica.
With some FSTs the number of arcs and symbols shrank considerably.


\section*{Acknowledgments}

I wish to thank Andreas Eisele (XRCE Grenoble)
and the anonymous reviewers of my article
for their valuable comments and advice.

\end{document}